\pdfoutput=1
\documentclass{mosi}

\usepackage{pifont}      %
\usepackage{enumitem}    %
\usepackage{float}       %
\usepackage{makecell}    %
\usepackage{tabularx}    %

\graphicspath{{assets/}{figures/}{./}}


\definecolor{oursgray}{gray}{0.95}   %

\newcommand{\bnum}[1]{\textbf{#1}}   %
\newcommand{\snum}[1]{\underline{#1}} %
\newcommand{\cmark}{\ding{51}}
\newcommand{\xmark}{\textcolor{black!30}{\ding{55}}}
\newcommand{\statetok}[1]{\texttt{<|#1|>}}

\newcommand{\mossvl}{MOSS-VL}
\newcommand{\mossbase}{MOSS-VL-Base}
\newcommand{\mossinstruct}{MOSS-VL-Instruct}
\newcommand{\mossrealtime}{MOSS-VL-Realtime}

\newtcolorbox{promptbox}[2][]{
  colback=white, coltext=black,
  arc=3mm, boxrule=0.5pt, colframe=black!60!white,
  title={#2}, colbacktitle=black, coltitle=white, fonttitle=\bfseries,
  top=8pt, bottom=8pt, left=10pt, right=10pt,
  breakable,
  before upper={\linespread{1}\selectfont
    \setlength{\parskip}{1ex plus 0.2ex minus 0.2ex}\setlength{\parindent}{0pt}},
  #1
}

\title{MOSS-VL Technical Report}

\author{OpenMOSS Team$^{*}$}
\authornote{$^{*}$Full contributors are listed in the
Contributors section.}

\abstract{%
We present \mossvl{}, an open vision--language model family that treats
real-time interaction---perceiving while it speaks---as a first-class
capability. It is co-designed across the stack: the language decoder
attends to vision only through gated cross-attention, so the model
can naturally see incoming frames while generating; a synthesized
interaction corpus
supervises when to speak, when to stay silent, and when to revise; and
a staged curriculum concentrates all real-time-specific training in one
light final stage over a strong offline foundation. Offline,
\mossinstruct{} is competitive at comparable scale and leads
temporal-reasoning video sets. Across four
streaming benchmarks, \mossrealtime{} posts the best average on three
(second on the fourth) among open-source streaming models, sweeping the
three subsets that squarely test proactive behavior---66.0 vs.\ 37.5
for the best baseline on OmniMMI Proactive Alerting. With 11.3B
parameters but visual tokens outside
the decoded sequence, \mossvl{} widens its time-to-first-token
advantage over same-backbone Qwen3-VL-8B from $2.8\times$ to
$5.1\times$ as visual context grows. We release all five checkpoints,
the training curriculum, and the real-time inference code at
\url{https://github.com/OpenMOSS/MOSS-VL}.%
}

\checkdata[Project]{\url{https://openmoss.ai/MOSS-VL/}}
\checkdata[Code]{\url{https://github.com/OpenMOSS/MOSS-VL}}
\checkdata[Models]{\url{https://huggingface.co/OpenMOSS-Team}}
\checkdata[Contact]{\texttt{pywang24@m.fudan.edu.cn}, \texttt{xpqiu@fudan.edu.cn}}

\begin{document}
\maketitle

\section{Introduction}\label{sec:introduction}

\begin{figure}[b!]
  \centering
  \includegraphics[width=\linewidth]{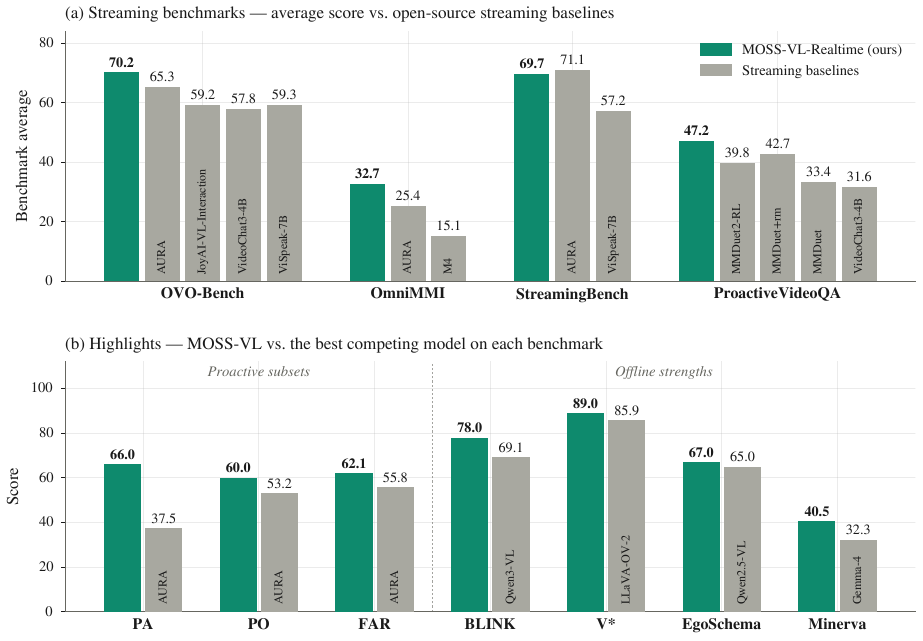}
  \caption{Results overview. (a) Average scores on four streaming
  benchmarks against open-source streaming baselines (details in
  \Cref{tab:streaming}; the StreamingBench average covers its visual
  groups, \S\ref{sec:eval-streaming}). (b) \mossvl{} against the best competing model on
  the proactive streaming subsets---PA (Proactive Alerting, OmniMMI),
  PO (Proactive Output, StreamingBench), and FAR (Forward Active
  Responding, OVO-Bench), all testing whether the model speaks up
  unprompted at the right moment---and on selected offline strengths
  (details in \Cref{tab:streaming,tab:offline}). Green = \mossvl{}; gray =
  the labeled competitor.}
  \label{fig:overview}
\end{figure}

\begin{table}[t]
  \centering
  \caption{Levels of video understanding, from offline to real-time. Each
  level lights up one additional capability axis; the dividing line between
  the streaming regime (L2--L4) and real-time (L5) is whether the model
  keeps perceiving \emph{while} it generates---L2--L4 models are blind
  during a reply, an L5 model is not. \mossrealtime{} achieves L5 behavior,
  demonstrated through live demos and the released real-time inference
  code, and is quantitatively validated at levels L2--L4 on four streaming
  benchmarks; a dedicated benchmark for L5 behavior remains an open
  problem.}
  \label{tab:levels}
  \small
  \renewcommand{\arraystretch}{1.3}
  \begin{tabular}{@{}clp{5.4cm}cccc@{}}
    \toprule
    & & & \multicolumn{4}{c}{Capability axes} \\
    \cmidrule(lr){4-7}
    Level & Regime & Defining behavior
      & \makecell{Stream.\\input}
      & \makecell{Silence}
      & \makecell{Multi-\\resp.}
      & \makecell{Perceive\\while gen.} \\
    \midrule
    L1 & offline & Watches the full video first, then answers questions
         about it (multi-turn allowed). & \xmark & -- & \xmark & \xmark \\
    L2 & streaming & Video arrives continuously; the user may ask at any
         moment and the model answers immediately---but it is blind while
         replying. & \cmark & \xmark & \xmark & \xmark \\
    L3 & streaming & Adds \emph{waiting}: if the answer is not yet
         determinable, the model stays silent until the key evidence
         appears, then answers. & \cmark & \cmark & \xmark & \xmark \\
    L4 & streaming & Adds \emph{persistent queries}: one question stays
         resident and the model re-answers as the scene evolves (still
         blind during each individual reply). & \cmark & \cmark & \cmark
         & \xmark \\
    L5 & real-time & Adds \emph{perception during generation}: the model
         keeps watching while it speaks, revising or interrupting its own
         reply the moment the evidence changes. & \cmark & \cmark & \cmark
         & \cmark \\
    \bottomrule
  \end{tabular}
\end{table}

Most open vision--language models understand video offline: given a
finished clip, they read it end to end and then answer questions about
it \citep{Bai2025qwen,An2026llava,Team2026gemma}. The settings where
video understanding matters most do not wait for the clip to end. A
live assistant watches a scene that is still unfolding, decides for
itself when something is worth saying, and must keep watching while it
says it. Table~\ref{tab:levels} organizes this capability space into
five levels. L1 is the offline regime. L2--L4 form the streaming
regime occupied by recent streaming models
\citep{Wang2024videollm,Fu2025vispeak,Xia2025streaming,Lu2026aura}:
input arrives continuously, deliberate silence and persistent queries
come into play, yet the model stays blind for the duration of each
reply. L5 adds the ability that separates real-time interaction from
everything below: perceiving \emph{while} generating, so a reply can be
revised or cut short the moment the evidence changes.

\mossvl{} is an open vision--language model family that treats
real-time interaction as a first-class capability, and it reaches that
capability by co-design rather than through any single component. The
architecture enables the behavior: the language decoder attends to
vision only through gated cross-attention, so visual tokens never enter
the decoded sequence, and an arriving frame merely appends to the
cross-attention cache---the model naturally keeps perceiving while it
generates (\S\ref{sec:architecture}). XRoPE orders text and vision along one
shared timeline, and absolute timestamp tokens make wall-clock time
explicit. The data injects the behavior: a synthesized corpus of
real-time interaction supervises when to speak, when to stay silent,
and how to revise a reply the scene has overturned
(\S\ref{sec:posttraining}). The training strategy keeps the stack
stable: a four-stage pre-training curriculum and standard supervised
fine-tuning build the offline foundation (\S\ref{sec:pretraining}),
every real-time-specific choice is concentrated in Realtime-SFT, one
light final stage, and a single system prompt moves the same weights
among offline, streaming, and real-time operation. \mossvl{} continues
a line that began with MOSS-Video-Preview \citep{Wang2026moss}, which
explored the real-time paradigm; this release redesigns the stack from
scratch and gives the line its first quantitative streaming evaluation.

Figure~\ref{fig:overview} previews the outcome. Offline,
\mossinstruct{} is competitive with open models of comparable scale
and leads the temporal-reasoning video sets Minerva, TOMATO, and
VideoMME-Logical (\S\ref{sec:eval-offline}). In the streaming regime,
the wins land precisely where timing is being tested: across four
streaming benchmarks against open-source streaming baselines,
\mossrealtime{} posts the best average on three of the four (and is
second on the fourth), sweeping the three subsets that squarely test
proactive behavior---66.0 vs.\ 37.5 on OmniMMI's Proactive Alerting
(\S\ref{sec:eval-streaming}). Efficiency follows from the same design:
against Qwen3-VL-8B, built on the same Qwen3-8B language backbone, the
time-to-first-token gap widens from $2.8\times$ to $5.1\times$ as
visual context grows (\S\ref{sec:eval-efficiency}). L5 behavior itself
is demonstrated qualitatively, through live demos and the released
real-time inference code (\S\ref{sec:eval-showcase}); quantitative
validation covers L2--L4, where public benchmarks exist.

\pagebreak
In summary, our main contributions are:
\begin{itemize}[leftmargin=1.5em, itemsep=2pt, topsep=2pt]
  \item \textbf{A stack co-designed for real-time interaction.} The
  architecture enables it---gated cross-attention with XRoPE and
  absolute timestamps lets the model take in new frames naturally
  while it generates;
  the data injects it---synthesized streams supervise response
  timing; the training strategy keeps it stable---the language backbone
  stays intact behind zero-initialized gates.
  \item \textbf{Realtime-SFT, an interaction paradigm in one light
  stage.} Two added state tokens, one shared system prompt, and a
  reweighted next-token loss---under 3\% of total training
  tokens---teach when to speak, when to stay silent, and when to
  revise, delivering the best average on three of four streaming
  benchmarks and the proactive sweep above.
  \item \textbf{Timing understanding shows up offline as well.}
  \mossinstruct{}, trained without the real-time corpus, leads the
  temporal-reasoning video sets in our
  comparison---Minerva, TOMATO, VideoMME-Logical---echoing the
  perception and temporal-understanding foundations laid during
  pre-training.
  \item \textbf{More parameters, faster serving.} \mossvl{} spends 11.3B
  parameters, yet visual tokens stay outside the decoded sequence, so
  serving latency grows more slowly with visual context than its
  same-backbone interleaved counterpart---measured in SGLang, not
  estimated. We release the complete training curriculum and all five
  checkpoints.
\end{itemize}

\section{Architecture}\label{sec:architecture}

\begin{figure}[t]
  \centering
  \includegraphics[width=\linewidth]{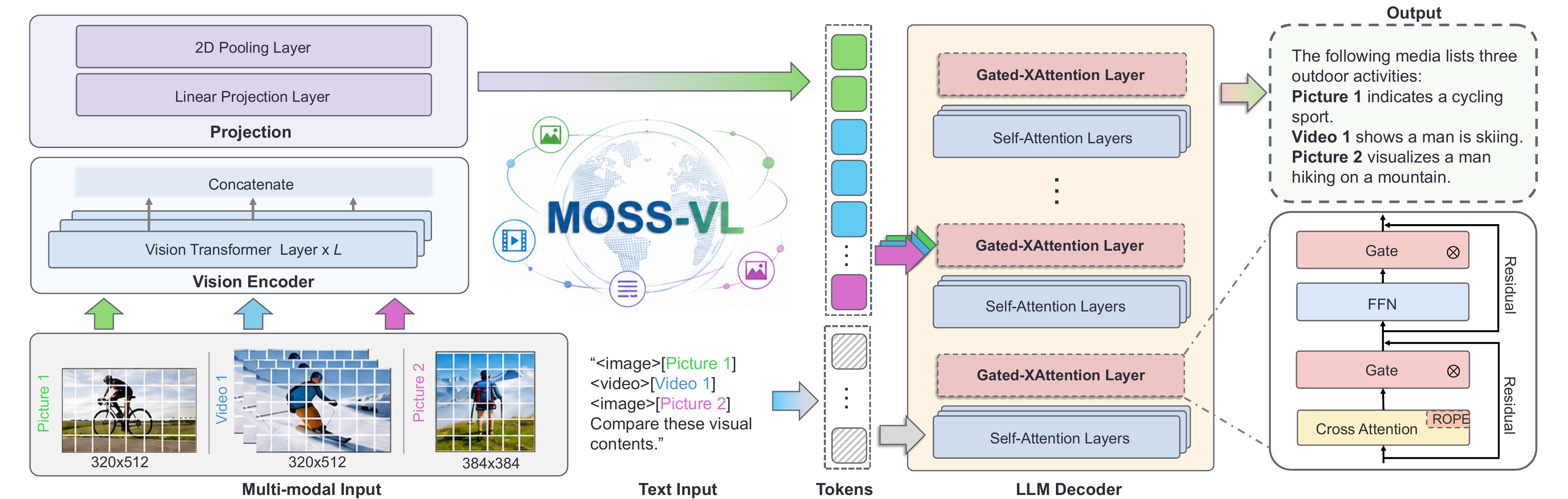}
  \caption{\mossvl{} architecture. Images and video frames are encoded by
  a 27-layer vision encoder at native dynamic resolution, then pooled
  $2\times2$ and projected into visual tokens; text is tokenized alongside.
  The LLM decoder (48 layers, initialized from Qwen3-8B) attends to visual
  tokens only through 12 tanh-gated cross-attention layers
  (Gated-XAttention, right inset), one in every four; the other 36
  self-attention layers operate on the text sequence alone, so visual
  tokens never join the decoded sequence. Cross-attention queries carry
  text positions and keys carry three-axis XRoPE coordinates $(t,h,w)$.
  Gates are zero-initialized scalars, so training starts from an intact
  language backbone.}
  \label{fig:arch}
\end{figure}

\begin{figure}[t]
  \centering
  \includegraphics[width=\linewidth]{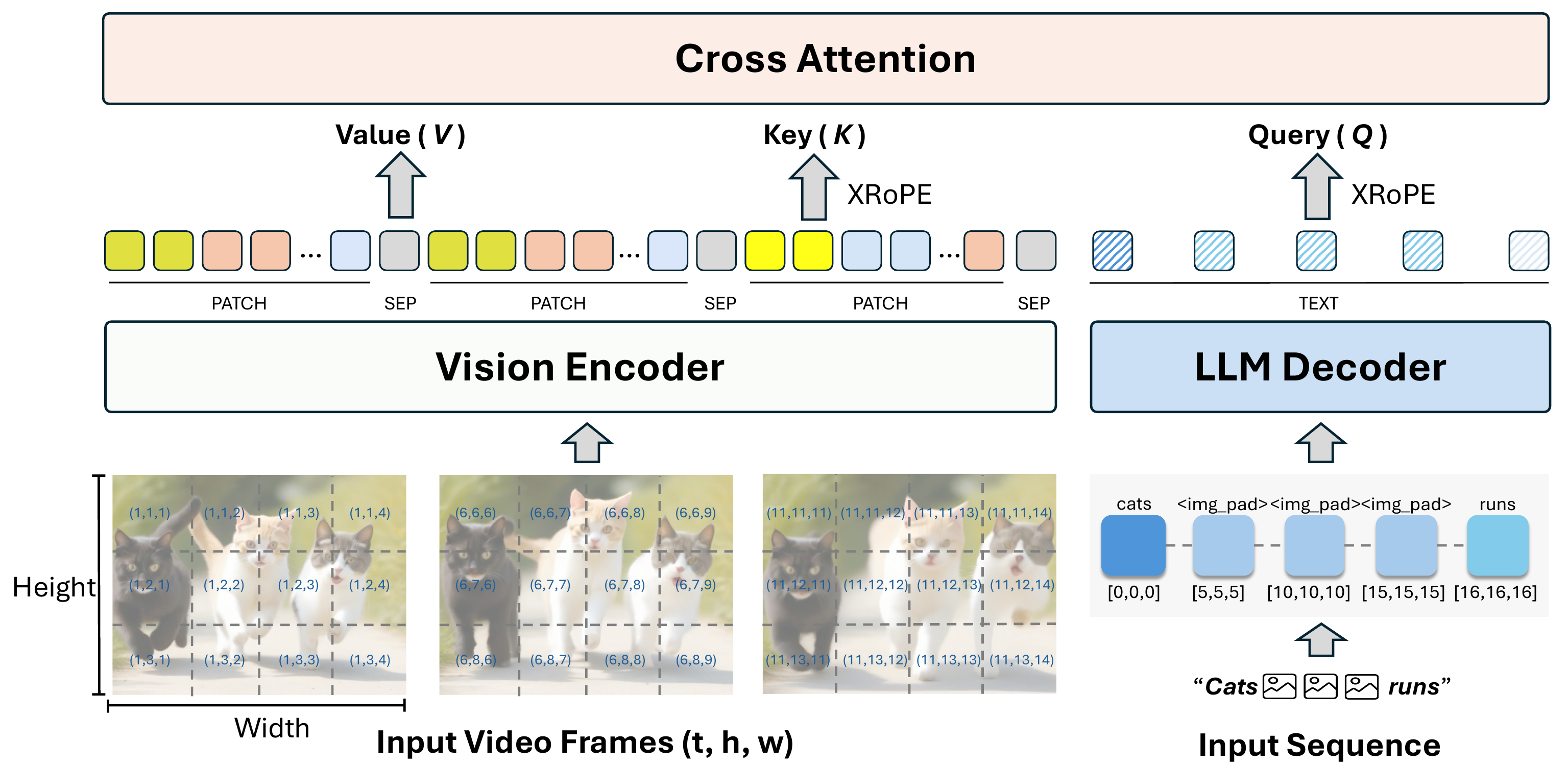}
  \caption{XRoPE, the position encoding of the cross-attention channel.
  Text tokens and visual patches share one three-axis coordinate space
  $(t,h,w)$, ordered by their logical position in the stream: a text
  token advances all three axes together, while a frame anchored at
  coordinate $t$ tiles its patches from $(t,t,t)$ at the top-left to
  $(t,t{+}h{-}1,t{+}w{-}1)$ at the bottom-right. The separator token
  closing each frame on the vision side and that frame's placeholder
  token in the text stream receive the same coordinate, so both channels
  advance along a single shared timeline. Rotations are applied to
  text-side queries and vision-side keys before they meet in
  cross-attention.}
  \label{fig:xrope}
\end{figure}

\begin{table}[t]
  \centering
  \caption{\mossvl{} configuration, identical across all released
  checkpoints. Video-input ranges reflect the native training setup; the
  released processor defaults to 1\,fps and 256 frames, and our
  evaluations keep 1\,fps with the cap raised to 768 frames
  (\S\ref{sec:evaluation}).}
  \label{tab:config}
  \small
\setlength{\tabcolsep}{6pt}
\renewcommand{\arraystretch}{1.15}
\begin{tabular}{@{}ll@{}}
\toprule
\multicolumn{2}{@{}l}{\itshape Language decoder (initialized from Qwen3-8B)} \\
\midrule
Layers            & 48 = 36 self-attention + 12 gated cross-attention \\
Cross-attention placement & every 4th layer (indices $2, 6, \ldots, 46$) \\
Hidden size / FFN size    & 4096 / 12288 \\
Attention heads   & 32 query / 8 key--value (GQA), head dim.\ 128 \\
Position encoding & XRoPE: interleaved 3-axis RoPE $(t,h,w)$, \\
                  & \quad sections $(24,20,20)$, base $5\times10^{6}$ \\
Context window    & 262{,}144 tokens \\
Vocabulary        & 151{,}936 \\
\midrule
\multicolumn{2}{@{}l}{\itshape Vision encoder (initialized from Qwen3-VL)} \\
\midrule
Layers            & 27 \\
Hidden size / FFN size & 1152 / 4304 \\
Attention heads   & 16 \\
Patch size        & $16\times16$ spatial, 1 frame temporal \\
Feature levels    & layers $\{8, 16, 24\}$ + final layer \\
Projection        & $2\times2$ spatial merge + MLP $\rightarrow$ 4096 \\
Input resolution  & native dynamic, 4096--16.8M pixels \\
\midrule
\multicolumn{2}{@{}l}{\itshape Video input} \\
\midrule
Frame sampling    & dynamic 1--16 fps, motion-adaptive (1--2 fps typical) \\
Frames per video  & up to 2{,}048 in training; 768 in our evaluations
                    (released default 256) \\
\midrule
Total parameters  & 11.3B (BF16) \\
\bottomrule
\end{tabular}

\end{table}

\begin{table}[t]
  \centering
  \caption{Released \mossvl{} checkpoints. The 0708 run is the subject of
  this report: \mossbase{} is the pre-trained model, \mossinstruct{} is its
  instruction-tuned successor, and \mossrealtime{} continues training from
  \mossinstruct{} with Realtime-SFT. The 0408 checkpoints are an earlier,
  independently trained run of the same architecture, released alongside
  for research continuity. All five are available under the OpenMOSS-Team
  organization on HuggingFace.}
  \label{tab:lineage}
  \small
\setlength{\tabcolsep}{5pt}
\renewcommand{\arraystretch}{1.15}
\begin{tabular}{@{}llll@{}}
\toprule
Checkpoint & Initialized from & Training & Evaluated in \\
\midrule
MOSS-VL-Base-0708     & --- & pre-training (\S\ref{sec:pretraining}) & --- \\
MOSS-VL-Instruct-0708 & Base-0708 & SFT (\S\ref{sec:posttraining}) & offline (\Cref{tab:offline}) \\
MOSS-VL-Realtime      & Instruct-0708 & Realtime-SFT (\S\ref{sec:posttraining}) & streaming (\Cref{tab:streaming}) \\
\addlinespace
MOSS-VL-Base-0408     & --- & pre-training & --- \\
MOSS-VL-Instruct-0408 & Base-0408 & SFT & --- \\
\bottomrule
\end{tabular}

\end{table}

\mossvl{} pairs a native-resolution vision encoder with a language
decoder initialized from Qwen3-8B \citep{Yang2025qwen}, and the two
interact only through gated cross-attention (Figure~\ref{fig:arch}).
Visual tokens never enter the decoded sequence: each frame contributes
a few timestamp tokens and one placeholder token to the text stream,
while its patch tokens are consumed as cross-attention keys and values.
Table~\ref{tab:config} lists the configuration.

\subsection{Components and Parameter Accounting}\label{sec:arch-components}

The vision encoder is a 27-layer transformer initialized from the
Qwen3-VL vision encoder \citep{Bai2025qwen}; it processes images and
frames at native resolution, from 4{,}096 to 16.8M pixels, drawing
features from three intermediate layers and the final layer. The
projection module merges each $2\times2$ patch group and maps it into
the decoder's hidden space. The decoder stacks 48 layers: 36
self-attention layers carried over from Qwen3-8B, and 12 gated
cross-attention layers, one at every fourth position. Of the 11.3B
total parameters, roughly 8.2B form the language backbone, 2.3B the
cross-attention stack, and 0.8B the vision encoder and projection
module.

\subsection{Gated Cross-Attention}\label{sec:xattn}

Each cross-attention layer follows the gated design of Flamingo
\citep{Alayrac2022flamingo}---queries come from the text hidden
states, keys and values from the visual tokens---implemented here
with grouped-query attention (32 query / 8 key--value heads) and
QK-RMSNorm. The layer wraps its
attention and feed-forward paths in tanh gates whose scalars are
zero-initialized, so training starts from an intact language backbone
(\S\ref{sec:pretraining}). The 36 self-attention layers never see
visual tokens.

\subsection{XRoPE}\label{sec:xrope}

To our knowledge, XRoPE (cross-attention rotary position embedding) is
the first position encoding introduced for the cross-attention channel
of a vision--language architecture: it gives the visual stream position
information that this channel otherwise lacks. XRoPE places text tokens
and visual patches in one three-axis coordinate space $(t,h,w)$,
ordered by their logical position in the stream
(Figure~\ref{fig:xrope}). A text token advances all three axes
together, taking coordinate $(x,x,x)$. A frame whose merged patch grid
is $h'\times w'$ anchors at the coordinate $t$ following the preceding
text, and its patches tile
\begin{equation}\label{eq:xrope}
  \mathbf{p}_{a,b} = (t,\; t+a,\; t+b),
  \qquad 0 \le a < h',\;\; 0 \le b < w',
\end{equation}
so height and width offsets ride on the shared temporal anchor. The
separator token that closes the frame on the vision side and the
frame's placeholder token in the text stream both take $(x,x,x)$ with
$x = \max(t+h',\, t+w')$, and the next text token continues from
$x+1$: the two channels advance along a single timeline. The 64 rotary
frequency pairs are split $(24,20,20)$ across $(t,h,w)$, and rotations
are applied to text-side queries and vision-side keys before they meet
in cross-attention. The $t$ axis is a relative sequence coordinate,
not wall-clock time; real timing enters through the timestamp tokens
of \S\ref{sec:timestamps}.

\subsection{Absolute Timestamps}\label{sec:timestamps}

Positions alone say nothing about wall-clock time, and frame rates
vary: \mossvl{} samples video at 1--16\,fps, motion-adaptive
(Table~\ref{tab:config}). Each frame is therefore preceded in the text
stream by an absolute timestamp,
\statetok{time\_start}\texttt{X.X seconds}\statetok{time\_end}, so
the model reads real time from tokens rather than inferring it from
positions, and timing stays explicit under any sampling rate.

\subsection{Real-Time by Construction}\label{sec:realtime-arch}

When a new frame arrives, only that frame is encoded; its keys and
values are appended to the cross-attention cache, and earlier frames
are neither re-encoded nor their keys and values recomputed. The
decoded sequence grows by the frame's
timestamp tokens and a single placeholder token---patch tokens stay on
the vision side---so an arriving stream leaves the decoding state
intact, and the next generated token already attends to the updated
cache through the gated layers.
\S\ref{sec:evaluation} quantifies the efficiency this
yields at inference time (Figure~\ref{fig:efficiency}).

\subsection{Released Models}\label{sec:released-models}

We release five checkpoints of one architecture
(Table~\ref{tab:lineage}). The 0708 run---\mossbase{},
\mossinstruct{}, \mossrealtime{}---is the subject of this report; the
0408 pair is an earlier, independently trained run of the same
architecture, released for research continuity.

\section{Pre-Training}\label{sec:pretraining}

\begin{table}[t]
  \centering
  \caption{The \mossvl{} training curriculum: four pre-training stages
  followed by SFT and Realtime-SFT (\S\ref{sec:posttraining}). The data
  of each stage is described in the corresponding subsection. Token
  counts are the tokens fed to the language model, with vision tokens
  counted after the $2{\times}2$ token compression of the projection
  module (\S\ref{sec:architecture}); sequence lengths are in tokens.}
  \label{tab:stages}
  \small
\setlength{\tabcolsep}{6pt}
\renewcommand{\arraystretch}{1.15}
\begin{tabular}{@{}lrrrcc@{}}
\toprule
Stage & Tokens & Samples & Max seq. & Trainable & Peak LR \\
\midrule
\multicolumn{6}{@{}l}{\itshape Pre-training} \\
\quad 1\enspace Vision--language alignment & 150.3B & 219.4M & 8K   & Projection + cross-attn & $2{\times}10^{-4}$ \\
\quad 2\enspace Large-scale multimodal     & 203.0B & 139.3M & 64K  & Full model & $5{\times}10^{-5}$ \\
\quad 3\enspace High-quality multimodal    & 459.0B &  14.7M & 128K & Full model & $1{\times}10^{-5}$ \\
\quad 4\enspace Annealing \& long-context  & 450.1B &   9.3M & 256K & Full model & $1{\times}10^{-5}$ \\
\midrule
\multicolumn{6}{@{}l}{\itshape Post-training (\S\ref{sec:posttraining})} \\
\quad SFT          &  102.8B & 7.6M  & 128K & Full model & $1{\times}10^{-5}$ \\
\quad Realtime-SFT &   34.8B & 0.56M & 256K & Full model & $4{\times}10^{-5}$ \\
\bottomrule
\end{tabular}

\end{table}

\mossvl{} is pre-trained with a four-stage curriculum: vision--language
alignment, large-scale multimodal pre-training, high-quality multimodal
pre-training, and a final stage of annealing and long-context training.
Table~\ref{tab:stages} lists the token budget, sample count, maximum
sequence length, trainable modules, and peak learning rate of every
stage, including the two post-training stages of
\S\ref{sec:posttraining}. The table shows the shape of the
curriculum: the maximum sequence length grows from 8K to 256K tokens,
and the token budget shifts toward the later stages while sample counts
fall by orders of magnitude---many short samples early, far fewer but
much longer and denser ones late. Throughout, the data is
decontaminated against our evaluation suites.

A defining trait of this training run is the scale of our data
synthesis. Alongside data collected and reorganized from existing
corpora, we synthesize high-quality caption, OCR, grounding, and
temporal-grounding data at large scale throughout the curriculum. These
four types cover the perceptual fundamentals of a vision--language
model---describing scenes, reading embedded text, localizing objects,
and anchoring events in time---and this synthesized core underpins the
strong perception and temporal-understanding foundations of the
released models.

\subsection{Stage 1: Vision--Language Alignment}\label{sec:pt-align}

Stage 1 connects the two pre-trained components. Only the newly
introduced parameters---the projection module and the cross-attention
layers---are updated, while the vision encoder and the language model
remain frozen; the high peak learning rate in Table~\ref{tab:stages}
applies to these fresh modules alone. The data comprises two
categories, image captioning and OCR, and sequences stay short at 8K
tokens.

\subsection{Stage 2: Large-Scale Multimodal Pre-Training}\label{sec:pt-scale}

With the connectors aligned, Stage 2 unfreezes the full model and
supplies breadth. The mixture spans image and video captioning, OCR,
grounding, interleaved image--text documents, and text-only
pre-training corpora, together with multimodal understanding data over
single images, multi-image sets, videos, and plain text across diverse
domains, and reasoning data. The context window extends to 64K tokens,
which admits long interleaved documents and video.

\subsection{Stage 3: High-Quality Multimodal Pre-Training}\label{sec:pt-quality}

Stage 3 spends the largest token budget of the curriculum
(Table~\ref{tab:stages}) on its highest-quality data. The mixture keeps
the Stage-2 categories but rebalances them: captioning recedes,
multimodal understanding and reasoning data take a larger share, and
mathematics, knowledge-intensive data, and temporal grounding enter the
mixture. Sequences extend to 128K tokens.

\subsection{Stage 4: Annealing and Long-Context Training}\label{sec:pt-anneal}

The final stage combines long-context training with high-quality
annealing. One data strand consists of long-video captioning,
long-video QA, and long-video temporal grounding, together with
long-document and long-text data, mixed with a small share of the
regular categories, and stretches sequences to 256K tokens. The other is an annealing mixture that re-weights
toward mathematics, knowledge-intensive data, and instruction-tuning
and QA data, and includes identity data. The curriculum yields
\mossbase{}, the starting point for post-training
(\S\ref{sec:posttraining}).

\section{Post-Training}\label{sec:posttraining}

Post-training proceeds in two supervised stages (Table~\ref{tab:stages});
neither uses reinforcement learning or a thinking mode. Standard
supervised fine-tuning (SFT) turns \mossbase{} into \mossinstruct{}, an
offline instruction follower. Realtime-SFT then continues from
\mossinstruct{} (Table~\ref{tab:lineage}) and installs the real-time
interaction paradigm: deciding at every frame whether to speak, staying
silent while nothing needs saying, and revising an answer when the scene
overturns it. Every real-time-specific design choice in \mossvl{} lives
in this final stage, which accounts for under 3\% of the total training
tokens.

\subsection{Supervised Fine-Tuning}\label{sec:sft}

We fine-tune \mossbase{} on 7.6M instruction samples (102.8B tokens)
with the standard next-token cross-entropy loss over assistant
responses, with sequences up to 128K tokens (Table~\ref{tab:stages}).
The samples combine data collected and reorganized from existing
corpora with data synthesized in house, and all of it passes filtering,
deduplication, decontamination against our evaluation suites, and
quality screening before entering the mixture. The
mixture covers general question answering over single images,
multi-image sets, videos, and plain text; perception-centric tasks
including OCR, document understanding, and spatial and temporal
grounding; image and video captioning; and reasoning-centric tasks
spanning multimodal reasoning, mathematics and other academic
disciplines, code, and knowledge-intensive QA, together with identity
data.

\subsection{Realtime-SFT: Learning When to Speak}\label{sec:realtime-sft}

Realtime-SFT teaches the model to treat incoming video as a stream of
decisions rather than a finished artifact. Training samples interleave
text with frames in arrival order: every frame is followed by a decision
slot, and each slot takes one of three forms---\statetok{silence} (keep
watching), \statetok{response} followed by text (speak now), or a reply
that ends with \statetok{silence} (finish speaking). A reply is spread
over consecutive slots frame by frame, emulating rate-limited real-time
output, and a single user turn may contain several separate emissions.
Supporting this costs exactly two new vocabulary entries---the two state
tokens, initialized from the embeddings of semantically related existing
tokens. The speak-or-wait decision itself is ordinary next-token
prediction: whenever the most probable next token is \statetok{silence},
the model waits for the next frame; otherwise it decodes a reply. No
dedicated decision head is attached.

\paragraph{Data characteristics.}
What sets the Realtime-SFT corpus (0.56M samples, ${\approx}$34.8B
tokens; Table~\ref{tab:stages}) apart is that every sample casts the
model in an explicit interaction role rather than a plain QA role:
standing instructions that must fire exactly once when their condition
is met; resident questions whose answers must update as evidence
accumulates; continuous real-time commentary; counting that accumulates
across a stream; probes of whether this is the right moment to speak;
and video-independent dialogue that maintains identity consistency. A
further share of offline QA and general multimodal data preserves
offline ability.

\paragraph{Data construction.}
The corpus draws on two sources. We first collect open-source datasets
for streaming video understanding and subject them to strict filtering
and re-annotation. More important, however, is the data we synthesize
ourselves, targeting the behaviors that existing datasets provide
scarcely or not at all: staying silent until evidence appears, revising
an answer as the scene evolves, and recovering when a new event
interrupts a reply midway. Synthesis follows the caption-driven pipeline
introduced in MOSS-Video-Preview \citep{Wang2026moss}: hierarchical,
densely time-anchored captions are mined for state transitions of a
focal object; each transition yields a question, an immediately
answerable reply, and a trajectory of updates; replies are anchored to
visual moments, laid out over frames with silence in between, and
filtered for quality. This round upgrades three of the pipeline's four
stages. Temporal anchoring is now verified frame by frame against the
actual footage: each reply is assigned the moment its evidence becomes
visible and the moment it stops being valid. Hand-offs are more
natural: a reply overtaken by a new event is rewritten as a
plain-language self-correction instead of being marked with a dedicated
interrupt token. Finally, a quality gate checks every sample against
its frames and keeps only those whose replies are grounded in what is
visible at emission time.

\paragraph{Corpus statistics.}
The interaction-first emphasis is visible in the numbers. Across the
corpus the model is supervised on 2.2M emission decisions, 58.7\% of
which are self-timed rather than prompted by a fresh user question; in
5.1\% of samples the target event never occurs, and the correct
behavior is to stay silent throughout. Streams run at 1\,fps for up to
768 frames (${\approx}$12.8 minutes) per window. The corpus is
decontaminated against our evaluation suites: benchmark videos are
excluded via held-out lists.

\subsection{Mode Control and Training Objective}\label{sec:mode-control}

Mode control in \mossvl{} amounts to a single system prompt. The
streaming and real-time modes share one prompt---real-time operation is
a special case of streaming---and offline inference uses none. One set
of weights thus operates in three inference modes with zero architecture
change. The prompt is reproduced below; the full dialogue template,
including frame and timestamp interleaving, is given in
Appendix~\ref{app:template}.

\begin{promptbox}{The shared streaming / real-time system prompt}
You are a helpful AI assistant specializing in real-time video analysis.
The video streams to you frame by frame. At every frame, you decide
independently whether to respond or stay silent --- output
\statetok{silence} when nothing relevant has happened, and respond when
the visual content warrants it.
\end{promptbox}

Supervision covers only what the assistant controls: reply text and the
two state tokens. System and user turns and the expanded visual tokens
are excluded from the loss. The central difficulty is imbalance: silence
slots vastly outnumber emission decisions, and under uniform weights the
model simply learns to stay silent. We therefore reweight the two state
tokens with a focal factor and inverse-frequency class coefficients,
computing the class statistics over the global batch at every step,
which keeps the class coefficients identical across data-parallel
ranks:
\begin{equation}\label{eq:realtime-loss}
\mathcal{L}=\frac{\sum_i m_i\, w_i\, \ell_i}{\sum_i m_i},\qquad
w_i=\begin{cases}
\alpha_{y_i}\,(1-p_i)^{\gamma} & \text{$y_i$ is a state token,}\\
1 & \text{otherwise,}
\end{cases}
\end{equation}
where $\ell_i$ is the token-level cross-entropy, $m_i$ the supervision
mask, $p_i$ the predicted probability of the target token, and
$\gamma=2$. The coefficient $\alpha_k=(n_s+n_r)/(2\,n_k)$, with $n_s$ and
$n_r$ counting silence and response targets in the current global batch,
equalizes the nominal, pre-focal weight of the two classes; reply text
keeps unit weight. Before focal modulation, the decision to speak and
the decision to stay silent thus carry equal aggregate weight in the
state-token loss, no matter how rare speaking is.

One further masking choice matters specifically in streams. In offline
chat, the token closing an assistant turn marks the end of an exchange;
in a stream, a turn boundary usually means the user interjected while the
world---and the conversation---continue. We therefore also exclude the
assistant's turn-final end token from supervision, so the model never
learns to wrap up merely because a new user turn appears. In a controlled
single-variable comparison, this masking raised emission frequency by
39\% and mean reply length by 68\%.

The behaviors installed here are evaluated quantitatively on four
streaming benchmarks in \S\ref{sec:evaluation} and qualitatively in
Figure~\ref{fig:showcase}.

\section{Infrastructure}\label{sec:infrastructure}

\mossvl{} is trained on a Megatron-LM stack \citep{Shoeybi2019megatron}
that combines data, tensor, sequence, and context parallelism to carry
the curriculum from 8K- to 256K-token sequences
(Table~\ref{tab:stages}). Variable-length multimodal samples are packed
into full sequences, keeping batches dense across mixed image, video,
and text data.

\paragraph{FlashAttention for cross-attention.}
The gated cross-attention of \S\ref{sec:xattn} has a visibility pattern
that off-the-shelf attention kernels do not serve: each text query
attends to the visual tokens of every frame that precedes it in the
stream (\S\ref{sec:realtime-arch}). Visibility is thus a per-query
prefix of the key--value sequence, growing frame by frame.
FlashAttention exposes causal or windowed masks, and materializing the
pattern as a dense cross-attention mask instead costs memory and
bandwidth proportional to the product of the text and visual sequence
lengths. We therefore extend FlashAttention-3
\citep{Shah2024flashattention} with a compact interface,
\texttt{cross\_kv\_boundary}, which encodes the visible prefix of each
query row as one 32-bit integer and carries it through the operator
schema, the scheduler, and the CUDA forward and backward kernels.
Key--value tiles beyond a row's boundary are pruned rather than
computed and masked, so kernel cost tracks visibility. The backend
covers the dense, variable-length, and KV-cache execution
paths---which keeps it compatible with packed training---and is
released with the model as a derivative of the upstream implementation.

\begin{figure}[t]
  \centering
  \includegraphics[width=\linewidth]{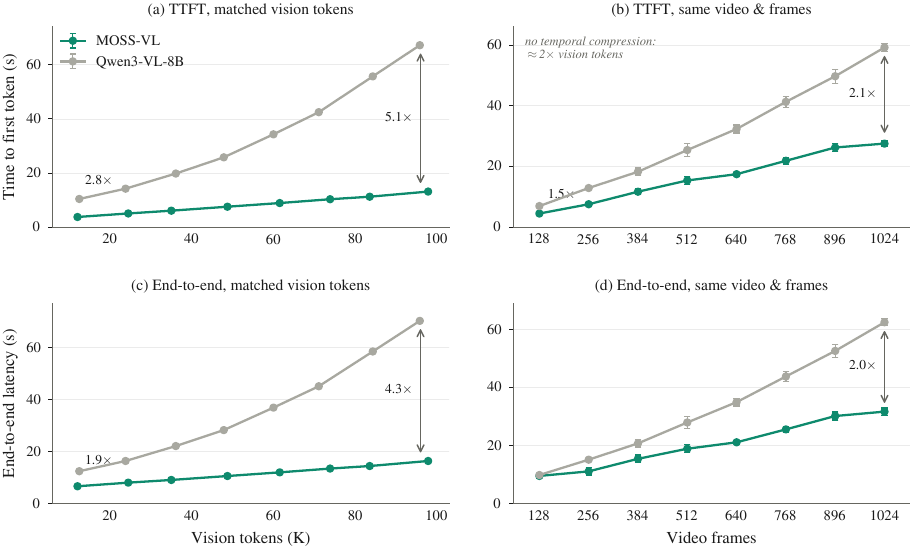}
  \caption{Measured serving latency of \mossvl{} vs.\ Qwen3-VL-8B --- the
  same Qwen3-8B language backbone, isolating the vision-integration
  architecture. Both models serve offline with SGLang on a single H200
  (TP=1, BF16, identical engine version) with an identical
  generation-length cap, which every run reaches; every point is the mean
  of five independent cold starts (error bars: sample std). (a,\,c) With ViT
  output matched, the time-to-first-token (TTFT) gap widens from $2.8\times$ to
  $5.1\times$ as visual context grows, and end-to-end latency from
  $1.9\times$ to $4.3\times$. (b,\,d) On the same video
  at the same resolution and frame count: Qwen3-VL compresses the
  temporal axis $2\times$, whereas \mossvl{} forgoes temporal compression
  so that each arriving frame can be encoded immediately --- carrying
  about twice the vision tokens on the same input --- yet it never
  falls behind, and its lead grows with stream length. Slower latency
  growth follows from the append-only
  cross-attention design: visual tokens never enter the decoded sequence
  (\S\ref{sec:architecture}).}
  \label{fig:efficiency}
\end{figure}

\paragraph{Serving and release.}
Our SGLang \citep{Zheng2023sglang} integration of \mossvl{} is merged
upstream, and the offline-serving measurements of
\S\ref{sec:evaluation} (Figure~\ref{fig:efficiency}) run on this stack.
Real-time interaction ships separately as a Transformers reference
implementation, released with the model weights on GitHub and
HuggingFace.

\section{Evaluation}\label{sec:evaluation}

\begin{table}[!t]
  \centering
  \caption{Offline results: \mossinstruct{} (0708 release) against
  open-source models of comparable scale, blocked by capability domain.
  Bold = best, underline = second best, ``--'' = not reported;
  LLaVA-OV-2 = LLaVA-OneVision-2-8B. Evaluation protocol and baseline
  provenance are spelled out in \S\ref{sec:eval-offline}.}
  \label{tab:offline}
  \footnotesize
\setlength{\tabcolsep}{4.5pt}
\renewcommand{\arraystretch}{1.12}
\begin{tabular}{@{}lccccc@{}}
\toprule
Benchmark & \makecell{\textbf{MOSS-VL}\\\textbf{(ours)}} & \makecell{Qwen3-VL\\8B} & \makecell{Qwen2.5-VL\\7B} & \makecell{LLaVA-OV-2\\8B} & \makecell{Gemma-4\\12B-IT} \\
\midrule
\multicolumn{6}{@{}l}{\itshape Multimodal perception} \\
MMBench-EN (v1.1)~\citep{Liu2023mmbench} & \bnum{88.1} & 84.8 & 83.2 & \snum{85.8} & 82.7 \\
MMStar~\citep{Chen2024right} & 66.0 & \snum{70.9} & 63.9 & 64.8 & \bnum{74.9} \\
RealWorldQA       & 68.0 & \bnum{71.5} & 68.5 & \snum{69.7} & 65.6 \\
MME-RealWorld~\citep{Zhang2024realworld} & \bnum{66.3} & -- & \snum{57.4} & -- & 46.9 \\
BLINK~\citep{Fu2024blink} & \bnum{78.0} & \snum{69.1} & 56.4 & 63.5 & 65.6 \\
POPE~\citep{Li2023evaluating} & \bnum{89.4} & -- & \snum{87.4} & -- & 81.4 \\
MMMU (val)~\citep{Yue2023mmmu} & 51.1 & \snum{69.6} & 58.6 & -- & \bnum{69.7} \\
CountBench~\citep{Paiss2023teaching} & 85.9 & 80.5 & -- & \snum{89.0} & \bnum{90.2} \\
CVBench~\citep{Tong2024cambrian} & 85.7 & \snum{86.7} & -- & \bnum{87.7} & 84.6 \\
V*~\citep{Wu2023guided} & \bnum{89.0} & 85.3 & -- & \snum{85.9} & 51.8 \\
MuirBench~\citep{Wang2024muirbench} & 39.9 & \bnum{64.4} & 59.6 & -- & \snum{61.6} \\
AI2D~\citep{Kembhavi2016diagram} & 81.9 & \bnum{85.7} & 83.9 & 84.3 & \snum{84.8} \\
\midrule
\multicolumn{6}{@{}l}{\itshape Video understanding} \\
VideoMME~\citep{Fu2024video} & 68.1 & \snum{71.4} & 65.1 & \bnum{71.9} & 60.5 \\
VideoMME-v2~\citep{Fu2026video} & \bnum{12.7} & \snum{12.4} & 10.3 & -- & -- \\
VideoMME-v2 (sub)~\citep{Fu2026video} & \snum{16.5} & \bnum{18.2} & -- & -- & -- \\
VideoMME-Logical~\citep{Kwan2026video} & \bnum{17.1} & \snum{11.9} & 7.4 & -- & 10.8 \\
MLVU (dev)~\citep{Zhou2024mlvu} & \snum{76.8} & \bnum{78.1} & 70.2 & 76.6 & -- \\
LongVideoBench~\citep{Wu2024longvideobench} & 65.9 & \bnum{68.0} & 56.0 & \snum{66.9} & 58.2 \\
LVBench~\citep{Wang2024lvbench} & 51.1 & \bnum{58.0} & 45.3 & \snum{55.5} & 37.3 \\
EgoSchema (sub)~\citep{Mangalam2023egoschema} & \bnum{67.0} & -- & \snum{65.0} & -- & 62.2 \\
MVBench~\citep{Li2023mvbench} & 66.7 & \snum{68.7} & \bnum{69.6} & 66.2 & -- \\
VSI-Bench~\citep{Yang2024thinking} & \snum{62.2} & 59.4 & 28.3 & \bnum{70.9} & 25.9 \\
Minerva~\citep{Nagrani2025minerva} & \bnum{40.5} & -- & -- & -- & \snum{32.3} \\
TimeLens-Charades~\citep{Zhang2025timelens} & 51.5 & \bnum{56.0} & 43.6 & \snum{53.5} & -- \\
TimeLens-ANet~\citep{Zhang2025timelens} & \snum{49.1} & 46.8 & 31.4 & \bnum{53.8} & -- \\
TimeLens-QVH~\citep{Zhang2025timelens} & \snum{60.0} & 59.4 & 31.6 & \bnum{66.4} & -- \\
TOMATO~\citep{Shangguan2024tomato} & \bnum{39.5} & \snum{34.6} & -- & -- & 31.9 \\
\midrule
\multicolumn{6}{@{}l}{\itshape Grounding} \\
RefCOCO-REC~\citep{Yu2016modeling} & 84.4 & \bnum{91.6} & \snum{90.0} & -- & -- \\
Ref-Adv~\citep{Akula2020words} & \bnum{57.0} & 47.2 & \snum{49.3} & -- & -- \\
\midrule
\multicolumn{6}{@{}l}{\itshape Document / OCR} \\
DocVQA (val)~\citep{Mathew2020docvqa} & 89.6 & \bnum{96.1} & \snum{95.7} & 95.2 & 80.8 \\
ChartQA~\citep{Masry2022chartqa} & \snum{87.8} & \bnum{89.6} & 87.3 & 85.9 & 51.2 \\
InfoVQA (val)~\citep{Mathew2021infographicvqa} & 68.9 & \bnum{83.4} & \snum{82.6} & 74.4 & 52.0 \\
OCRBench~\citep{Liu2023ocrbench} & 86.1 & \bnum{89.6} & \snum{86.4} & 78.2 & 76.9 \\
OCRBench-v2~\citep{Fu2024ocrbench} & \snum{57.4} & \bnum{63.3} & 56.8 & -- & 39.6 \\
OmniDocBench (v1.6)~\citep{Ouyang2024omnidocbench} & \bnum{88.9} & \snum{84.9} & -- & -- & -- \\
\midrule
\multicolumn{6}{@{}l}{\itshape Reasoning} \\
VLMsAreBlind~\citep{Rahmanzadehgervi2024vision} & 63.5 & \snum{74.0} & -- & -- & \bnum{75.9} \\
VisuLogic~\citep{Xu2025visulogic} & \bnum{27.5} & 22.5 & \snum{26.0} & -- & -- \\
ERQA~\citep{Team2025gemini} & \bnum{45.8} & \bnum{45.8} & -- & \snum{43.3} & 40.8 \\
EmbSpatial~\citep{Du2024embspatial} & 70.7 & \bnum{78.5} & -- & \snum{78.1} & 72.5 \\
\bottomrule
\end{tabular}

\end{table}

We evaluate each model in the regime it is built for: \mossinstruct{}
on an offline suite of 39 benchmarks across five capability domains
(Table~\ref{tab:offline}), and \mossrealtime{} on four streaming
benchmarks---OVO-Bench \citep{Li2025bench}, OmniMMI
\citep{Wang2025omnimmi}, StreamingBench \citep{Lin2024streamingbench},
and ProactiveVideoQA \citep{Wang2025proactivevideoqa}---which together
cover levels L2--L4 of the capability hierarchy in
Table~\ref{tab:levels}. All streaming evaluation is carried out in
streaming fashion: frames are fed as they would arrive, and every
model runs under its own streaming protocol. Measured serving
efficiency (\S\ref{sec:eval-efficiency}) and qualitative real-time
sessions (\S\ref{sec:eval-showcase}) complete the picture.

\subsection{Offline Results}\label{sec:eval-offline}

Table~\ref{tab:offline} compares \mossinstruct{} with open models of
comparable scale---Qwen3-VL-8B \citep{Bai2025qwen}, Qwen2.5-VL-7B
\citep{Bai2025qwenvl25}, LLaVA-OneVision-2-8B \citep{An2026llava}, and
Gemma-4-12B-IT \citep{Team2026gemma}---over multimodal perception, video
understanding, grounding, document/OCR, and reasoning.
\mossvl{} results sample video at 1\,fps with at most 768 frames,
following a benchmark's official protocol wherever one is prescribed;
DocVQA and InfoVQA use the validation split. Baseline numbers are
taken from the respective official reports and reflect their authors'
inference settings, which may differ from ours, particularly in video
frame count. The exceptions are the Gemma-4-12B-IT column, which we
evaluated ourselves under the same protocol as \mossvl{}, and the
Qwen3-VL entry on OmniDocBench (v1.6), evaluated with the Markdown
prompt from the official Qwen3-VL cookbook; the remaining OmniDocBench
baselines are omitted, as their official numbers are not
metric-comparable.

Perception is the strongest block: \mossinstruct{} takes five of the
twelve rows---MMBench-EN (88.1), POPE (89.4), V* (89.0), and both
MME-RealWorld (66.3) and BLINK (78.0) by 8.9-point margins. On video,
it leads the temporal-reasoning sets---Minerva (40.5),
TOMATO (39.5), and VideoMME-Logical (17.1), each by 4.9 points or
more---plus EgoSchema (67.0), consistent with the
perception and temporal-understanding foundations built in
\S\ref{sec:pretraining}. The wins extend across the remaining
domains: the adversarial Ref-Adv grounding set goes to \mossinstruct{}
by 7.7 points (57.0), OmniDocBench document parsing by 4.0 (88.9), and on
reasoning it takes VisuLogic (27.5) and shares the top ERQA score
(45.8). The main gaps sit in MMMU, document understanding, and
standard grounding, both referring (RefCOCO-REC) and temporal
(TimeLens); the first two we return to in \S\ref{sec:discussion}.

\subsection{Streaming Benchmarks}\label{sec:eval-streaming}

Table~\ref{tab:streaming} reports subset-level results on the four
streaming benchmarks. Baselines are open-source streaming
models---AURA \citep{Lu2026aura}, M4 (released with OmniMMI itself)
\citep{Wang2025omnimmi}, ROMA \citep{Tian2026roma},
JoyAI-VL-Interaction \citep{Yao2026joyai}, VideoChat3-4B
\citep{Li2026videochat3}, ViSpeak-7B \citep{Fu2025vispeak}, and the
MMDuet family
\citep{Wang2024videollm,Wang2025proactivevideoqa,Wang2025mmduet2}; the
panel follows each benchmark's published coverage and therefore differs
across the four. Baseline numbers are taken from the respective
official reports, except MMDuet's OmniMMI entry, which its own report
does not cover and is taken from the ROMA report; \mossrealtime{}
numbers are from our own evaluation under each benchmark's official
protocol. %
\mossrealtime{} runs an 8.2B language backbone, comparable to
the 7--8B-class baselines; its additional parameters lie outside the
decoded sequence, in the vision encoder and the cross-attention stack
(\S\ref{sec:architecture}).

\begin{table}[!t]
  \centering
  \caption{Streaming benchmark results: \mossrealtime{} against
  open-source streaming baselines at subset level. Bold = best,
  underline = second best; ``--'' = subset not reported in that
  model's official report. Subset abbreviations, each benchmark's Avg
  convention, and data provenance are spelled out in
  \S\ref{sec:eval-streaming}.}
  \label{tab:streaming}
  \footnotesize
\setlength{\tabcolsep}{6pt}
\renewcommand{\arraystretch}{1.12}
\begin{tabular}{@{}lccccc@{}}
\toprule
\multicolumn{6}{@{}l}{\itshape OVO-Bench~\citep{Li2025bench}} \\
Subset & \textbf{MOSS-VL-Realtime} & AURA & JoyAI-VL-Interaction & VideoChat3-4B & ViSpeak-7B \\
\midrule
FAR  & \bnum{62.1} & \snum{55.8} & -- & -- & 54.2 \\
BT   & \bnum{72.6} & \snum{60.4} & -- & -- & 57.5 \\
RTVP & \snum{75.9} & \bnum{79.8} & -- & -- & 66.3 \\
Avg  & \bnum{70.2} & \snum{65.3} & 59.2 & 57.8 & 59.3 \\
\midrule
\multicolumn{6}{@{}l}{\itshape OmniMMI~\citep{Wang2025omnimmi}} \\
Subset & \textbf{MOSS-VL-Realtime} & AURA & M4 & ROMA & MMDuet \\
\midrule
PA  & \bnum{66.0} & \snum{37.5} & 25.5 & \snum{37.5} & 22.0 \\
SG  & \snum{21.7} & \bnum{24.0} & 5.7 & -- & -- \\
MD  & \bnum{10.7} & \snum{7.7} & 1.7 & -- & -- \\
AP  & \bnum{33.5} & \snum{32.0} & \bnum{33.5} & -- & -- \\
SI  & \bnum{31.5} & \snum{26.0} & 9.0 & -- & -- \\
Avg & \bnum{32.7} & \snum{25.4} & 15.1 & -- & -- \\
\midrule
\multicolumn{6}{@{}l}{\itshape StreamingBench~\citep{Lin2024streamingbench}} \\
Subset & \textbf{MOSS-VL-Realtime} & AURA & VideoChat3-4B & ViSpeak-7B & \\
\midrule
PO  & \bnum{60.0} & \snum{53.2} & -- & 50.8 & \\
RT  & 82.9 & \bnum{83.2} & \snum{83.0} & 70.4 & \\
CTX & \snum{56.4} & \bnum{59.0} & -- & 43.9 & \\
SQA & \snum{50.4} & \bnum{57.2} & -- & 39.2 & \\
Avg (visual) & \snum{69.7} & \bnum{71.1} & -- & 57.2 & \\
\midrule
\multicolumn{6}{@{}l}{\itshape ProactiveVideoQA~\citep{Wang2025proactivevideoqa}} \\
Subset & \textbf{MOSS-VL-Realtime} & MMDuet2-RL & MMDuet+rm & MMDuet & VideoChat3-4B \\
\midrule
WEB & \bnum{55.4} & \snum{53.3} & 43.5 & 38.9 & 38.4 \\
EGO & \snum{47.8} & 33.6 & \bnum{52.2} & 46.0 & 28.1 \\
TV  & \bnum{50.2} & \snum{43.4} & 32.6 & 21.1 & 34.7 \\
VAD & \snum{35.3} & 28.9 & \bnum{42.5} & 27.4 & 25.1 \\
Avg & \bnum{47.2} & 39.8 & \snum{42.7} & 33.4 & 31.6 \\
\bottomrule
\end{tabular}

\end{table}

Each benchmark reports at subset level. OVO-Bench separates forward
active responding (FAR), backward tracing (BT), and real-time visual
perception (RTVP). OmniMMI covers proactive alerting (PA), dynamic
state grounding (SG), multi-turn dependency (MD), action prediction
(AP), and speaker identification (SI); its Avg is the mean of the five
subsets, shown only for models with all five reported. StreamingBench
groups real-time visual understanding (RT) and contextual
understanding (CTX), with proactive output (PO) and sequential QA
(SQA), subsets of CTX, listed separately for the proactive analysis;
Avg (visual) is the mean of the RT and CTX group scores---\mossvl{}
takes no audio input, so the audio-dependent Omni-Source group is not
evaluated and no official overall score is reported. ProactiveVideoQA
splits by video source into web (WEB), egocentric (EGO), and TV-series
(TV) video QA, plus video anomaly detection (VAD).

\mossrealtime{} posts the best average on three of the four
benchmarks---OVO-Bench (70.2 vs.\ 65.3 for the runner-up), OmniMMI
(32.7 vs.\ 25.4), and ProactiveVideoQA (47.2 vs.\ 42.7)---and is
second on StreamingBench's visual average (69.7 vs.\ AURA's 71.1).

The subset pattern says more than the averages. The three subsets that
squarely test proactive behavior---speaking unprompted, at the right
moment---all go to \mossrealtime{}: Proactive Alerting on OmniMMI
(66.0 vs.\ 37.5), Proactive Output on StreamingBench (60.0 vs.\ 53.2),
and Forward Active Responding on OVO-Bench (62.1 vs.\ 55.8).
ProactiveVideoQA, proactive in every subset, follows in aggregate, and
Backward Tracing on OVO-Bench (72.6 vs.\ 60.4) shows the accumulated
stream history staying usable. The wins concentrate where response
timing is the skill under test---the behavior Realtime-SFT
supervises directly (\S\ref{sec:realtime-sft}); where a subset reduces
to perception QA over the current scene, AURA keeps the edge.

\subsection{Inference Efficiency}\label{sec:eval-efficiency}

Figure~\ref{fig:efficiency} measures serving latency against
Qwen3-VL-8B, which shares the Qwen3-8B language backbone
\citep{Yang2025qwen}; the comparison therefore isolates the
vision-integration architecture. Both models run offline SGLang
serving on a single H200 (\S\ref{sec:infrastructure}). With ViT output
matched, the time-to-first-token gap widens from $2.8\times$ to
$5.1\times$ as visual context grows, and end-to-end latency from
$1.9\times$ to $4.3\times$. The same-video comparison is stricter for
us: \mossvl{} forgoes temporal compression so that each arriving frame
can be encoded immediately (\S\ref{sec:realtime-arch}), and thus
carries about twice the vision tokens of Qwen3-VL on identical
input---yet it never falls behind at any measured point. The advantage widens
with visual context, which is exactly the regime a real-time assistant
occupies: visual history accumulates by the minute while replies must
keep arriving on time.

\subsection{Qualitative Results}\label{sec:eval-showcase}

\begin{figure}[t]
  \centering
  \includegraphics[width=\linewidth]{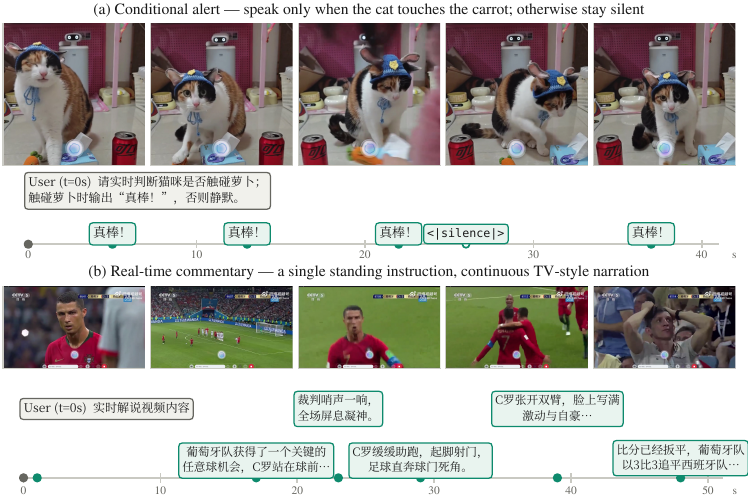}
  \caption{\mossrealtime{} in the wild: two screen-captured sessions from
  the released live demo, running on a single H200, shown as frame strips
  with excerpts of the model's outputs on a shared timeline (long outputs
  truncated with ``\ldots''; $t=0$ at the user instruction). (a) Under
  a standing conditional instruction (``say \emph{great!}\ whenever the
  cat touches the carrot, otherwise stay silent''), the model stays silent
  throughout and fires exactly at each of the four contacts; the hollow
  marker samples one of the silent spans, where the per-frame output is
  the silence token. (b) Under a
  single instruction (``commentate the video live''), commentary begins
  within a second and tracks a free-kick sequence --- set-up, the
  referee's whistle, the strike, the celebration, and the updated
  scoreline --- in a professional broadcast register. Outputs are in
  Chinese, the demo language; English translations of both sessions are
  given in Appendix~\ref{app:transcripts}, and these and further live
  cases can be viewed on the official blog
  (\url{https://openmoss.ai/MOSS-VL/}).}
  \label{fig:showcase}
\end{figure}

Figure~\ref{fig:showcase} shows the installed behaviors in live
operation, with two sessions from the released
demo. Under a standing conditional instruction, the model holds
silence over the full stream and fires at each of the four target
contacts, and only there; under a single commentary instruction, it
opens within a second and tracks a free-kick sequence through the
whistle, the strike, the celebration, and the updated scoreline.
Between them, the sessions exercise the core interaction roles of the
Realtime-SFT corpus (\S\ref{sec:realtime-sft})---a standing
instruction that must fire exactly when its condition is met, and
continuous real-time commentary---under real-world timing.

\section{Discussion}\label{sec:discussion}

Read as a whole, the evaluation shows a pattern rather than a score.
The streaming wins concentrate in the subsets that test \emph{when} to
speak; the serving advantage widens exactly where visual history
accumulates; the offline strengths cluster on temporal-reasoning
video sets. No single component explains this shape. It is what
co-design looks like from the outside: an architecture in which
perception runs naturally alongside generation, a corpus that supervises the
response timing, and a curriculum that builds the foundation before
one light final stage makes it interactive. We treat the full system,
not just the weights, as the release: alongside all five checkpoints come the staged
training curriculum (Table~\ref{tab:stages}), the complete Realtime-SFT
dialogue template (Appendix~\ref{app:template}), the real-time
inference implementation, the extended FlashAttention-3 backend
(\S\ref{sec:infrastructure}), and the recorded live sessions of
\S\ref{sec:eval-showcase}.

The limits are equally visible. \mossinstruct{} trails the strongest
open models of its scale on reasoning-heavy suites such as MMMU and on
document-centric benchmarks (Table~\ref{tab:offline}): \mossvl{} ships
without a thinking mode, and its training optimizes for real-time video
rather than exam-style reasoning. The capability this report is built
around is, at its highest level, still qualitatively attested:
quantitative validation stops at L2--L4 because public streaming
benchmarks stop there, and no existing benchmark measures perception
during generation---whether a model revises or cuts short a reply the
moment the scene overturns it.

Both limits mark the near-term agenda: reinforcement-learning
post-training for the \mossvl{} series, already on our public roadmap,
and a dedicated benchmark for L5 behavior, which the field still
lacks.

\section{Conclusion}\label{sec:conclusion}

\mossvl{} makes real-time interaction a first-class capability of an
open vision--language model family. It is built in, not bolted on:
gated cross-attention with XRoPE lets frames arrive while text is
being generated, synthesized interaction data teaches the model when
to speak, when to wait, and when to revise, and a staged curriculum
confines every real-time-specific choice to one light final stage. In
evaluation, \mossinstruct{} holds strong offline ground, especially
on temporal-reasoning tasks, \mossrealtime{} leads streaming
benchmarks wherever response timing is tested, and serving
latency grows more slowly with visual context than an interleaved
peer's.
Weights, curriculum, and code are open.

\section*{Contributors}\label{sec:contributions}

\noindent{\itshape Core Contributors}\par\nopagebreak\noindent
Pengyu Wang$^{*}$,
Chenkun Tan,
Shaojun Zhou,
Qirui Zhou,
Yanxin Chen,
Xingyang He,
Huazheng Zeng,
Jijun Cheng,
Chenghao Wang,
Xiaomeng Qian,
Pengfei Wang,
Zhan Huang,
Shanqing Gao,
Wei Huang,
Longjun Cao,
Wu Ran,
Jie Liu,
Changtai Zhu

\medskip
\noindent{\itshape Contributors}\par\nopagebreak\noindent
Hongkai Wang,
Yixian Tian,
Chenghao Liu,
Zhen Ye,
Xinghao Wang,
Botian Jiang,
Guoguo Feng,
Zhaoye Fei,
Ruixiao Li,
Mingshu Chen,
Yang Gao,
Qinyuan Cheng,
Shimin Li,
Xipeng Qiu$^{\S}$

\medskip
\noindent{\itshape Affiliations}\par\nopagebreak\noindent
Fudan University\\
Shanghai Innovation Institute\\
MOSI Intelligence

{\let\thefootnote\relax
\footnotetext{$^{*}$Project Lead.\quad$^{\S}$Corresponding Author.}}

\clearpage
\bibliographystyle{plainnat}
\bibliography{refs}

\clearpage
\appendix
\section{Real-Time Interaction Details}

\subsection{Real-Time Dialogue Template}\label{app:template}

All three inference modes share the released ChatML-style chat template.
Offline inference is standard: the full video is encoded as one vision
block and the model replies as an ordinary chat assistant, with no
dedicated system prompt. The streaming and real-time modes prepend the
shared system prompt of \S\ref{sec:mode-control} and lay every
assistant turn out as an alternating stream of decision slots $t_i$ and
frame placeholders,
\begin{center}
$t_0$\,\statetok{video}\,$t_1$\,\statetok{video}\,$t_2$%
\,$\cdots$\,\statetok{video}\,$t_N$,
\end{center}
so $N$ frames leave the model $N{+}1$ decisions. Each slot takes one of
three forms: \statetok{silence}---nothing to say at this frame;
\statetok{response} followed by a text chunk---speaking, not yet
finished; or a chunk closed by \statetok{silence}---the reply ends
here. The session below shows the layout end to end. Line breaks in the
actual byte stream occur only after each role header and after each
\texttt{<|im\_end|>}; every other break is wrapping, and the italic
annotations are not part of the stream.

\begin{promptbox}{Real-time session layout (template level)}
\ttfamily\footnotesize\hyphenpenalty=10000\exhyphenpenalty=10000\raggedright
<|im\_start|>system\\
\textrm{\itshape the shared streaming / real-time system prompt (\S\ref{sec:mode-control})}<|im\_end|>\\
\textrm{\itshape ---~warm-up: the stream is already flowing, nobody has spoken~---}\\
<|im\_start|>user\\
<|im\_end|>\\
<|im\_start|>assistant\\
<|silence|>\allowbreak<|video|>\allowbreak<|silence|>\allowbreak<|video|>\allowbreak<|silence|>\allowbreak\textrm{\itshape\ \dots\ nothing happens, the model keeps silent\ \dots}\allowbreak<|im\_end|>\\
\textrm{\itshape ---~a reply spread over consecutive frames; later the instruction fires a second time~---}\\
<|im\_start|>user\\
Tell me when the door opens.<|im\_end|>\\
<|im\_start|>assistant\\
<|silence|>\allowbreak<|video|>\allowbreak<|silence|>\allowbreak<|video|>\allowbreak<|silence|>\allowbreak<|video|>\allowbreak<|response|>The door is opening,\allowbreak<|video|>\allowbreak<|response|>and a man in a red jacket steps in.<|silence|>\allowbreak<|video|>\allowbreak<|silence|>\allowbreak<|video|>\allowbreak<|response|>The door opens again,\allowbreak<|video|>\allowbreak<|response|>another man, in a white shirt, steps in.<|silence|>\allowbreak<|im\_end|>\\
\textrm{\itshape ---~leading-slot dropout variant: the frame arrives first~---}\\
<|im\_start|>user\\
What is the second man up to?<|im\_end|>\\
<|im\_start|>assistant\\
<|video|>\allowbreak<|response|>He is sneaking around,\allowbreak<|video|>\allowbreak<|response|>and has not noticed the first man staring at him.<|silence|><|im\_end|>
\end{promptbox}

Five details of this layout carry the design:
\begin{enumerate}[leftmargin=1.5em]
\item \textbf{The warm-up turn.} A session may open before anyone has
  spoken: the user turn is empty, and the assistant answers every slot
  with \statetok{silence}. Silence-by-default is exercised before any
  instruction exists.
\item \textbf{One \statetok{response} per speaking frame.} A reply
  spread over $k$ frames carries $k$ \statetok{response} tokens, one
  ahead of each chunk---not a single token for the whole reply.
\item \textbf{Exactly one closing \statetok{silence} per reply}, placed
  in the same slot as the final chunk. It is the same vocabulary entry
  as the idle \statetok{silence}, distinguished only by position (after
  text: the reply is finished; alone after a frame: nothing to say),
  and it is the model's only end-of-reply signal.
\item \textbf{The leading slot is an augmentation site.} About half of
  the training samples keep the slot between the user's message and the
  next frame (second assistant turn above); the rest drop it (last
  turn). The model thus handles a question that lands together with a
  frame and one that lands a beat ahead of it equally well.
\item \textbf{A turn may contain several replies.} In the second
  assistant turn the model closes its reply, stays silent for a frame,
  then fires again unprompted when the door opens a second time---its
  standing instruction met anew, with no new user turn in between.
\end{enumerate}

The template keeps each frame as a single \statetok{video} placeholder;
the processor---the same code path at training and inference
time---expands it into a timestamped vision block:
\begin{center}
\texttt{<|vision\_start|><|time\_start|>7.0 seconds<|time\_end|>%
<|image\_pad|><|vision\_end|>}
\end{center}
The timestamp is plain text carrying the frame's arrival time, so the
model reads the current stream time directly from its input;
\statetok{image\_pad} is then replaced by the frame's visual tokens
according to its resolution. Training and inference must agree on this
expansion byte for byte---any deviation shifts the model off its
training distribution and suppresses emission.

At inference the runtime reproduces the training-time alternation with
a strict handshake: it pushes one frame, waits for the model to emit a
fresh \statetok{silence}, and only then pushes the next. While a long
reply is still unfolding, a length budget releases the next frame
instead, so the stream never stalls behind the reply. Replies therefore
unfold across consecutive slots, one chunk per frame, exactly as in
training (\S\ref{sec:realtime-sft}).

\subsection{Live-Demo Session Transcripts}\label{app:transcripts}

Figure~\ref{fig:showcase} shows excerpts of the model's Chinese outputs;
Table~\ref{tab:transcripts} gives our English translations of both
sessions, in output order.

\begin{table}[h]
  \centering
  \caption{English translations (ours) of the two live-demo sessions in
  Figure~\ref{fig:showcase}. The figure carries the Chinese excerpts;
  timestamps and the placement of each output on the stream are read off
  the figure's timeline.}
  \label{tab:transcripts}
  \small
  \renewcommand{\arraystretch}{1.25}
  \begin{tabular}{@{}lp{11.2cm}@{}}
    \toprule
    Turn & English translation \\
    \midrule
    \multicolumn{2}{@{}l}{\itshape (a) Conditional alert} \\
    User ($t{=}0$) & Watch in real time and judge whether the cat
      touches the carrot; output ``Great!''\ whenever it does, and stay
      silent otherwise. \\
    Model & ``Great!'', emitted at each of the four contacts; every
      other frame yields the silence token. \\
    \midrule
    \multicolumn{2}{@{}l}{\itshape (b) Real-time commentary} \\
    User ($t{=}0$) & Commentate on the video live. \\
    Model & Portugal win a crucial free-kick chance; Ronaldo stands
      over the ball\ldots \\
    Model & The referee's whistle sounds, and the whole stadium holds
      its breath. \\
    Model & Ronaldo takes a measured run-up and strikes---the ball
      flies straight for the corner of the goal. \\
    Model & Ronaldo spreads his arms wide, his face full of excitement
      and pride\ldots \\
    Model & The score is level---Portugal have pulled back to 3--3
      against Spain\ldots \\
    \bottomrule
  \end{tabular}
\end{table}

\end{document}